\pdfoutput=1

\documentclass[11pt]{article}

\usepackage{EMNLP2023}  

\usepackage{times}
\usepackage{latexsym}

\usepackage[T1]{fontenc}

\usepackage[utf8]{inputenc}

\usepackage{microtype}

\usepackage{inconsolata}

\usepackage{tcolorbox}
\usepackage{graphicx}
\usepackage{multirow}
\usepackage{booktabs}

\usepackage{pifont}       
\usepackage{bbding}       
\usepackage{fontawesome}  

\newcommand{\cmark}{\textcolor{green}{\ding{51}}}
\newcommand{\xmark}{\textcolor{red}{\ding{55}}}

\usepackage{xspace}
\newcommand{\othree}{\mbox{\textsc{o3}}\xspace}
\newcommand{\gpt}{\mbox{\textsc{o4-mini}}\xspace}
\newcommand{\gemini}{\mbox{\textsc{Gemini 2.5 Pro}}\xspace}
\newcommand{\grok}{\mbox{\textsc{Grok 3}}\xspace}
\newcommand{\mist}{\mbox{\textsc{Mistral}}\xspace}
\newcommand{\claude}{\mbox{\textsc{Claude 3.7 Sonnet}}\xspace}
\newcommand{\llava}{\mbox{\textsc{LLaVA-v1.5-7B}}\xspace}
\newcommand{\doubao}{\mbox{\textsc{Doubao-1.5-pro}}\xspace}
\newcommand{\kimi}{\mbox{\textsc{Kimi-VL-A3B-Thinking}}\xspace}
\newcommand{\qwx}{\mbox{\textsc{Qwen2-VL-7B-Instruct}}\xspace}
\newcommand{\qwy}{\mbox{\textsc{Qwen2-VL-72B-Instruct}}\xspace}
\newcommand{\deepseek}{\mbox{\textsc{DeepSeek-VL2}}\xspace}

\title{SemVink: Advancing VLMs' Semantic Understanding of Optical Illusions via Visual Global Thinking}

\author{Sifan Li\textsuperscript{1}\hspace{1.4em}Yujun Cai\textsuperscript{2}\thanks{~~Corresponding author.}\hspace{1.6em}Yiwei Wang\textsuperscript{1}
\\
\textsuperscript{1}University of California, Merced
\hspace{1em}
\textsuperscript{2}University of Queensland
\\
\texttt{{sflijohn@foxmail.com}}
\hspace{1em}
\texttt{{yujun.cai@uq.edu.au}}
\hspace{1em}
\texttt{{yiweiwang2@ucmerced.edu}}
\\
  {\href{https://johnnyzeppelin.github.io/vlm-semvink/}{\textcolor{magenta}{\texttt{https://johnnyzeppelin.github.io/vlm-semvink}}}}
  }

\begin{document}
\maketitle

\begin{abstract}
Vision-language models (VLMs) excel in semantic tasks but falter at a core human capability: detecting hidden content in optical illusions or AI-generated images through perceptual adjustments like zooming. We introduce HC-Bench, a benchmark of 112 images with hidden texts, objects, and illusions, revealing that leading VLMs achieve near-zero accuracy (0–5.36\%) even with explicit prompting. Humans resolve such ambiguities instinctively, yet VLMs fail due to an overreliance on high-level semantics. Strikingly, we propose SemVink (Semantic Visual Thinking) by simply scaling images to low resolutions, which unlocks over 99\% accuracy by eliminating redundant visual noise. This exposes a critical architectural flaw: VLMs prioritize abstract reasoning over low-level visual operations crucial for real-world robustness. Our work urges a shift toward hybrid models integrating multi-scale processing, bridging the gap between computational vision and human cognition for applications in medical imaging, security, and beyond.
\end{abstract}

\section{Introduction}
\label{sec:intro}


\begin{figure}[t]
    \centering
    \includegraphics[width=\linewidth]{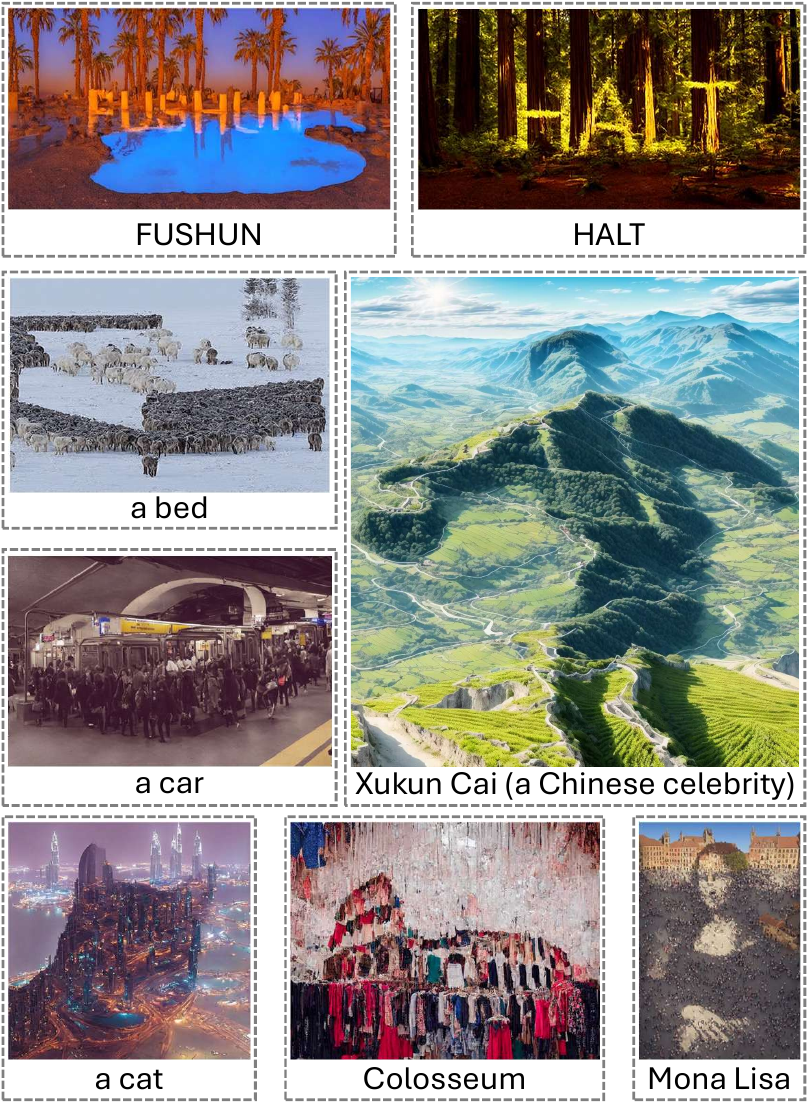}
    \caption{Illusional images can contain hidden texts or hidden images within the obvious background scenes.}
    \label{fig:dataset}
\end{figure}

Vision-language models (VLMs) have revolutionized multimodal understanding, excelling at tasks like image captioning and visual reasoning.
Although VLMs have been capable of many challenging visual tasks, some seemingly simple vision-language tasks are impossible for them to solve.
A critical gap persists: their inability to recognize visually hidden content, i.e., information embedded in images that requires human-like perceptual adaptations such as zooming, squinting, or dynamic scaling to detect. This limitation becomes starkly apparent when analyzing optical illusions, AI-generated ``double images,'' or medical scans with subtle anomalies, where human observers instinctively adjust their visual processing to uncover obscured details.

Current VLM architectures prioritize high-level semantic reasoning at the expense of low-level visual operations fundamental to human perception. While benchmarks like EXAMS-V~\citep{das2024examsv} test compositional reasoning, they neglect perceptual adaptability—the ability to iteratively refine visual analysis through multi-scale or contrast adjustments. This oversight masks a critical weakness: VLMs universally fail to detect hidden text or objects, even when explicitly prompted to ``zoom in'' or ``adjust contrast'', as shown in Figure~\ref{fig:problems}. The root cause lies in their reliance on static, high-resolution embeddings that prioritize local texture over global structure, burying hidden patterns under redundant spatial features.

We address this gap through three contributions. First, we introduce \textbf{HC-Bench}, a benchmark for hidden content recognition, comprising 112 synthetic images with embedded Latin/non-Latin texts and objects. Generated via the Stable Diffusion model~\citep{rombach2022} with ControlNet~\citep{controlnet}, these images preserve naturalistic backgrounds while embedding content detectable only through perceptual adjustments. Second, we demonstrate universal failure across 12 state-of-the-art VLMs (0–5.36\% accuracy), revealing their inability to simulate human-like visual refinement via prompting or few-shot learning. Third, we propose \textsc{\textbf{SemVink} (\textbf{Sem}antic \textbf{Vi}sual Thi\textbf{nk}ing)} to identify a surprisingly effective solution: programmatic image scaling to low resolutions (32–128 pixels), which suppresses redundant features and achieves over 99\% accuracy. Embedding analysis confirms that scaling shifts attention from local textures to global patterns, mirroring human perceptual strategies.

Our contributions are as follows:
\begin{itemize}
    \item {
    We introduce a benchmark for hidden content recognition, addressing limitations in existing datasets like EXAMS-V~\citep{das2024examsv} and IllusionBench~\citep{2025illusionbench}.
    }
    \item {
    We empirically reveal the VLMs’ inability to perform human-like perceptual adjustments, exposing a foundational design flaw prioritizing semantics over basic visual processing.
    }
    \item {
    We propose a scalable solution via preprocessing pipelines, demonstrating that low-level operations can bridge the gap between computational vision and human cognition.
    }
\end{itemize}

Our findings challenge the prevailing focus on semantic abstraction in VLM design. This study redefines VLM evaluation by emphasizing the importance of integrating low-level visual skills into multimodal architectures which is a paradigm shift critical for real-world robustness in ambiguous scenarios. By linking encoder limitations to redundant feature patterns, we provide actionable insights for improving VLM design.



\section{Related Work}
\label{sec:related}

Our research intersects three critical domains: (1) architectural limitations of vision-language models, (2) computational analysis of hidden content, and (3) multimodal benchmarking paradigms. We contextualize our contributions within these areas.

\subsection{Vision-Language Models}
\label{subsec:vlm}

Modern VLMs like CLIP~\citep{radford2021}, Flamingo~\citep{flamingo}, and BLIP-2 (Li et al., 2023) excel at semantic alignment between images and text, enabling tasks such as open-vocabulary detection and visual question answering. However, their design prioritizes high-level reasoning over low-level visual processing. Recent studies reveal critical gaps:
texture bias and static processing.
VLMs inherit CNNs’ tendency to prioritize local textures over global shapes~\citep{geirhos2022}, hindering recognition of content requiring spatial coherence~\citep{yang2024spatialcoherencelossobjects}.
Unlike humans, VLMs process images at fixed resolutions without dynamic scaling~\citep{dosovitskiy2021imageworth16x16words}, limiting adaptability to multi-scale patterns.
Redundant Embeddings: High-resolution vision encoders (e.g., ViT-L/14)\footnote{The model is available at \url{https://huggingface.co/openai/clip-vit-large-patch14}} produce spatially redundant features that obscure subtle details~\citep{llava,vasu2024fastvlmefficientvisionencoding,rao2024ravenmultitaskretrievalaugmented,carvalho-martins-2025-efficient}, corroborating our findings in Section~\ref{subsec:embedding}.

While recent work explores hybrid architectures~\citep{chen2024vitamindesigningscalablevision,qi2024roravlmrobustretrievalaugmentedvision,li2025surveystateartlarge,liao2025multimodalmambadecoderonlymultimodal} and multi-task training~\citep{rao2024ravenmultitaskretrievalaugmented,wang-etal-2023-efficientvlm,ma-etal-2024-modula}, none address perceptual adaptability for hidden content detection.

\subsection{Hidden Content and Perceptual Illusions}
\label{subsec:illu}

The study of hidden content spans cognitive science and computational vision.
Classic work on perceptual grouping~\citep{wertheimer-1923-group} and figure-ground segregation~\citep{Kanizsa} demonstrates humans’ ability to resolve ambiguous stimuli through iterative adjustments (e.g., squinting). Neuroimaging studies link this to feedback loops in visual cortex~\citep{LAMME2000571}.

With the advancement of generative AI, AI-generated images with hidden content emerge.
Diffusion models now embed text/objects imperceptible to humans without scaling~\citep{rombach2022}, raising concerns about adversarial misuse~\citep{10655785,zeng2025guarddoorsafeguardingmaliciousdiffusion,gao-etal-2024-empowering,duan2025}. ControlNet~\citep{controlnet} enables precise spatial conditioning but has not been leveraged for perceptual evaluation.
\citet{zhang-etal-2023-grounding} tested whether VLMs have the similar kind of illusions as humans do, which provides a new profound prospect to address this type of problems.
While IllusionBench~\citep{2025illusionbench} tests geometric illusions, and SVO-Probes~\citep{svo} evaluates spatial understanding, neither addresses AI-generated hidden content requiring dynamic processing.

\subsection{Multimodal Benchmarking Gaps}
\label{subsec:multimodal}

Existing benchmarks inadequately assess perceptual adaptability. We can find the three preference of existing benchmarks: focusing on semantic tests, robustness and dynamic processing, respectively.

VQA~\citep{vqa}, GQA~\citep{gqa}, and TextVQA~\citep{textvqa} emphasize textual or compositional reasoning, not low-level vision.

ImageNet-C~\citep{imagenetc} evaluates corruption resilience but not hidden content. EXAMS-V~\citep{das2024examsv} focuses on factual knowledge, not perceptual strategies.

\citet{multiscalevision} on multi-scale vision and \citet{neural2024} on neural compression highlight the need for adaptive resolution but lacks task-specific benchmarks.

\citet{hemmat2024hiddenplainsightevaluating}
focuses on abstract shape illusions and uses clearly defined geometric patterns without subtly hiding. This work only evaluates existing VLMs capabilities without proposing solutions. The broader problem of hidden-content recognition is not addressed yet.

\citet{rostamkhani2024illusoryvqabenchmarkingenhancing} introduced blurring method to address this recognition difficulty which has little effects.

SemVink fills these voids by comprehensively evaluating VLMs’ capacity to simulate human perceptual adjustments which is a prerequisite for robustness in real-world scenarios like medical imaging (subtle lesions) or security (steganographic content) and proposing a simple but effective solution to address this problem.

\subsection{Low-Level Vision Operations in Multimodal Models}
\label{subsec:lowlevel}

While recent VLM research emphasizes high-level alignment and reasoning, a complementary line of work shows that \emph{low-level} image operations can substantially improve robustness and perception. Classical multi-scale designs aggregate information across resolutions to suppress spurious textures and emphasize global structure (e.g., Multiscale Vision Transformers)~\citep{multiscalevision}. Closely related, anti-aliasing (low-pass filtering before downsampling) restores shift stability and reduces high-frequency artifacts that often dominate features in high-resolution encoders~\citep{zhang2019antialiased}.

Training-time transformations further indicate the utility of explicit, lightweight preprocessing. Corruption benchmarks such as ImageNet-C~\citep{imagenetc} spurred augmentation techniques (e.g., AugMix and RandAugment) that improve robustness to distribution shifts and nuisance variations without architectural changes~\citep{hendrycks2020augmix,cubuk2020randaugment}. Beyond distribution shift, the adversarial-robustness community has repeatedly found that \emph{input transformations} (bit-depth reduction, JPEG compression, total-variation minimization, smoothing) can ablate brittle, high-frequency cues and enhance resilience~\citep{guo2018inputtransform,xu2018featuresqueezing}.

SemVink operationalizes this perspective for VLMs: a simple \emph{zoom-out} (downscaling to 32--128 px) suppresses redundant high-frequency embeddings and exposes global patterns necessary for reading hidden text or recognizing silhouettes. This aligns with recent findings that many security-focused protections for modern generators and editors also rely on low-level transforms (e.g., watermark-embedded adversarial examples; protective backdoors) to manipulate or gate what models perceive and edit~\citep{10655785,zeng2025guarddoorsafeguardingmaliciousdiffusion}. We argue that such operations should be elevated from ad-hoc preprocessing to \emph{first-class, integrable visual tools} inside VLM pipelines (e.g., dynamic multi-scale routing or learned resolution schedules), complementing high-level semantics with the perceptual adaptability humans routinely employ.


\subsection{Cognitive Basis of Vision}
\label{rel:cognit}
Our work draws inspiration from theories of human vision perception. Some key theories are primary to both hidden content generation and recognition in our work.
Marr’s primal sketch that early visual processing extracts edges and blobs~\citep{marr}. This is analogous to our low-resolution scaling’s emphasis on global structure.
Predictive coding is also vital in human recognition.
Humans iteratively refine predictions through feedback~\citep{raoballard1999}, which is a capability absent in feedforward VLMs.
In perceptual learning, expertise improves detection of hidden patterns through reweighting visual features~\citep{goldstone1998}, suggesting potential for VLM fine-tuning with our proposed dataset HC-Bench.


\section{Methodology}
\label{sec:meth}

In this section, we introduce the dataset we construct and the zoom-out method we propose to help the models see the hidden content. The dataset is not only for our experiments but also for facilitating the future research in this topic.


\begin{figure*}[t]
    \centering
    \includegraphics[width=\linewidth]{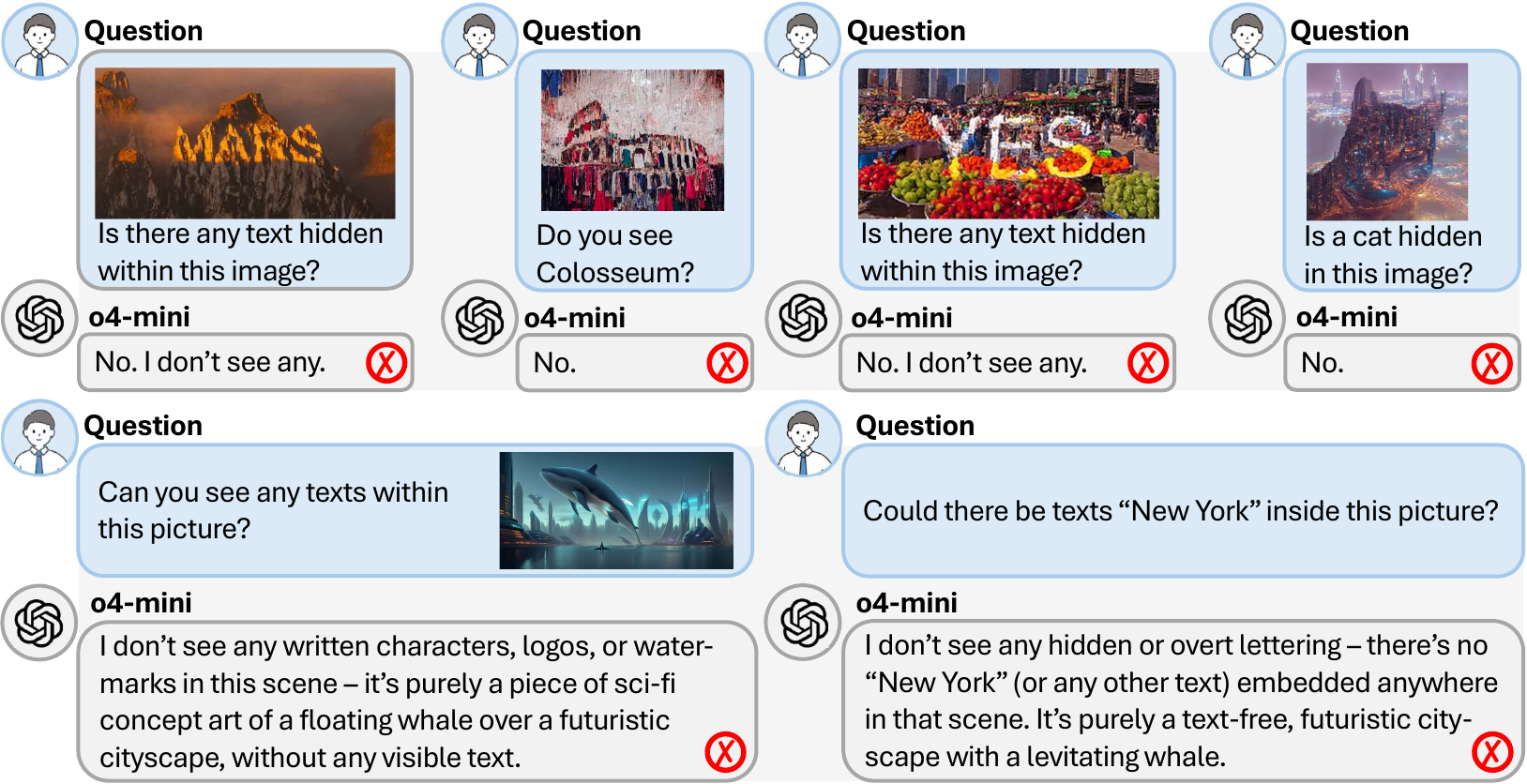}
    \caption{As one of the best state-of-the-art VLMs, \gpt is incapable in recognizing the hidden texts and objects within images even when we prompt directly with the correct answers. The hidden items in these images are \textit{``MARS'', Colosseum, ``YES'', a cat, and ``NEW YORK''}, respectively.}
    \label{fig:problems}
\end{figure*}

\subsection{Data Construction}
\label{subsec:data}

We introduce \textbf{HC-Bench}, a benchmark dataset for evaluating VLMs’ ability to recognize visually hidden content. As shown in Figure~\ref{fig:dataset}, the dataset consists of 112 synthetic images divided into two categories:

\paragraph{Hidden text images (56 total).} 28 Latin words are selected from seven semantic categories (e.g., animals, objects), varying in length and frequency. 28 non-Latin words are Chinese characters and other scripts, balanced for visual complexity.

\paragraph{Hidden object images (56 total).} Seven object classes (e.g., faces, animals, vehicles), with eight instances per class. Objects are subtly embedded into naturalistic backgrounds (e.g., landscapes, urban scenes).

We ensure the dataset is balanced to mitigate potential biases and enhance the generalizability.
For each type of concepts, we pick common concepts (e.g., words like \textit{``MARS''} and \textit{``DOG''} and objects like \textit{a cat} and \textit{a bed}) as half of the dataset and relatively rare concepts (e.g., words like \textit{``WYVERN''} and \textit{``SACCHARINE''} and objects like \textit{Cologne cathedral} and \textit{a Tyrannosaurus}) as the other half. The distribution is balanced as in Table~\ref{tab:dataset}.

\subsubsection{Implementation Details}
\label{dataset:details}

To hide the target content, we need a background scene that is irrelevant to what to hide.
We use Qwen3-235B-A22B\footnote{The model is available at \url{https://chat.qwen.ai/} and \url{https://huggingface.co/Qwen/Qwen3-235B-A22B}} to generate 112 diverse scene descriptions (e.g., \textit{a bustling city street at sunset} or \textit{a serene mountain lake}).

\begin{table}[ht]
\centering
\begin{tabular}{l | cc}
Type
& Hidden Text
& Hidden Object
\\
\hline
Normal
& 28 & 28
\\
Rare
& 28 & 28
\\
\end{tabular}
\caption{The data distribution of HC-Bench.}
\label{tab:dataset}
\end{table}

With these background scenes descriptions, we can hide the target content into the scene when synthesizing the image.
All images were generated using the diffusion model Stable Diffusion v1.5\footnote{The model is available ar \url{https://huggingface.co/stable-diffusion-v1-5/stable-diffusion-v1-5}}~\citep{rombach2022} with a specialized ControlNet~\citep{controlnet} module (control\_v1p\_sd15\_qrcode\_monster\footnote{The model is available at \url{https://huggingface.co/monster-labs/control_v1p_sd15_qrcode_monster}}) to ensure precise integration of hidden content.
We employ DPM++ 3M SDE~\citep{DPM} as the sampling method.
We set the ControlNet weight in the range from 1.2 to 1.4, since weights < 1.2 resulted in hidden content that is imperceptible for humans; weights > 1.4 caused unnatural artifacts.
The synthetic images are set to be with a resolution that either height or width is in the range of 512–1440 pixels (maintaining the aspect ratio).

The entire generation process is employed on one NVIDIA RTX A6000 card with 48 GB VRAM.



\begin{figure*}
    \centering
    \includegraphics[width=\linewidth]{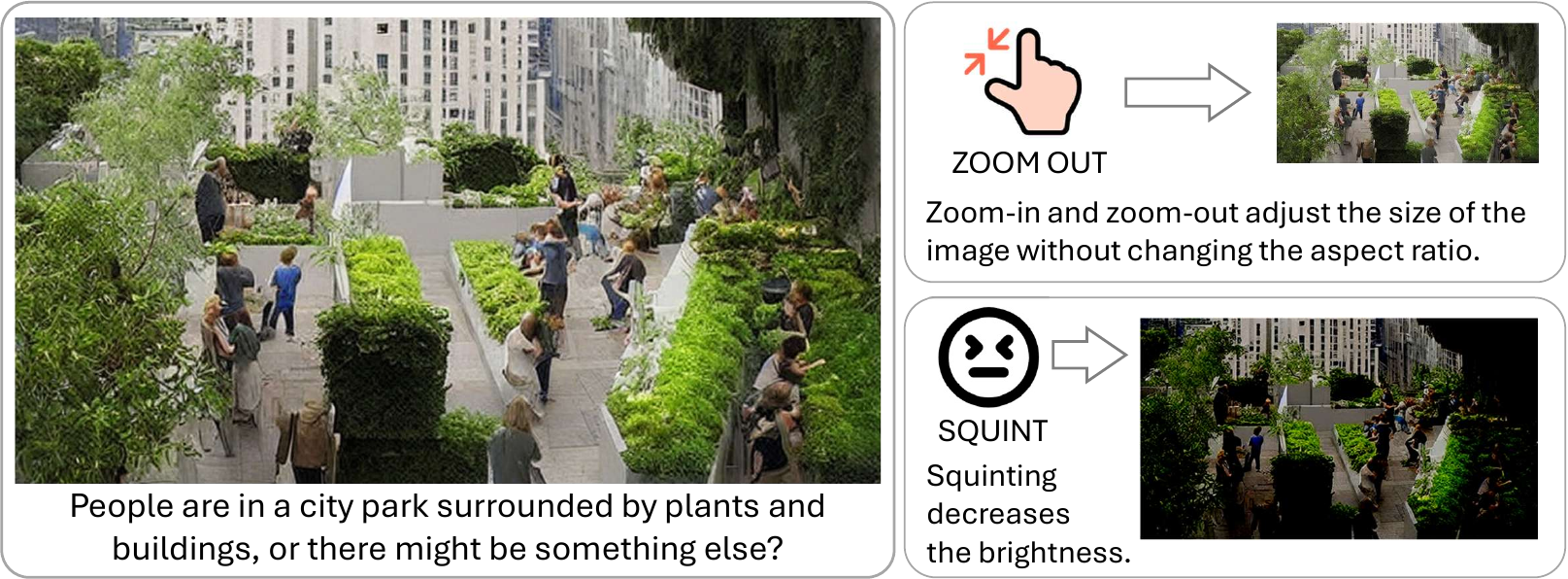}
    \caption{Two methods to help humans recognize the hidden content \textit{a Labrador retriever} within the image: zoom out the image to a sight from a distance and squint to observe the image to reduce the brightness to highlight the hidden content.}
    \label{fig:methods}
\end{figure*}

\subsection{Evaluation Protocol}
\label{method:zeroshot}

As shown in Figure~\ref{fig:problems}, we should ask VLMs with direct questions and then hint them with correct answers if direct questions do not obtain positive responses.

\paragraph{Direct questions.}
We first ask direct questions to VLMs. For hidden text cases, we ask:

\begin{tcolorbox}[fonttitle = \small\bfseries, title=Direct Question for Hidden Text,colframe=gray!2!black,colback=gray!2!white,boxrule=1pt,boxsep=0pt,left=5pt,right=5pt,fontupper=\footnotesize, halign title = flush center]
\begin{minipage}{.39\linewidth}
\includegraphics[width=\linewidth]{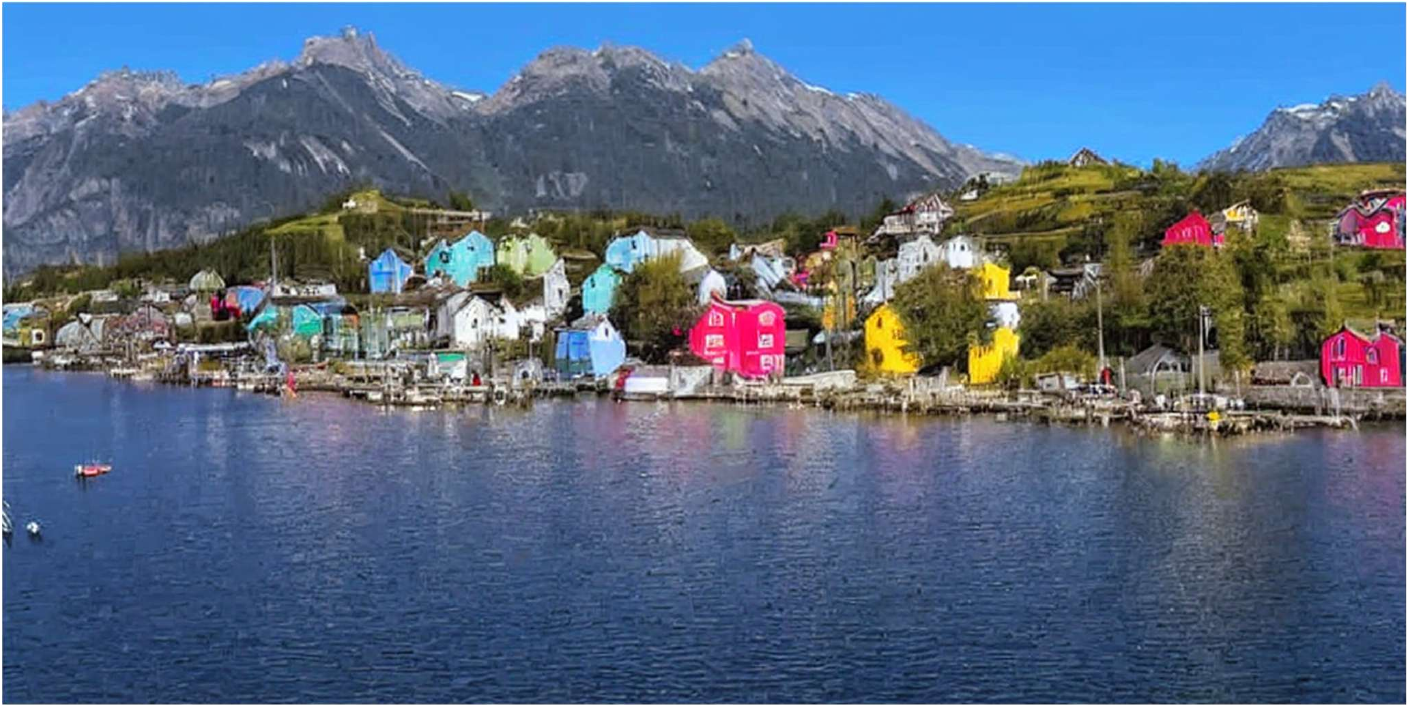}
\end{minipage}
\hspace{1em}
\begin{minipage}{.5\linewidth}
What is within this image? Is there any text hidden within this image?

\textcolor{blue}{[The hidden text: ``POLO'']}
\end{minipage}
\end{tcolorbox}

For hidden object cases, we ask:

\begin{tcolorbox}[fonttitle = \small\bfseries, title=Direct Question for Hidden Object,colframe=gray!2!black,colback=gray!2!white,boxrule=1pt,boxsep=0pt,left=5pt,right=5pt,fontupper=\footnotesize, halign title = flush center]
\begin{minipage}{.19\linewidth}
\includegraphics[width=\linewidth]{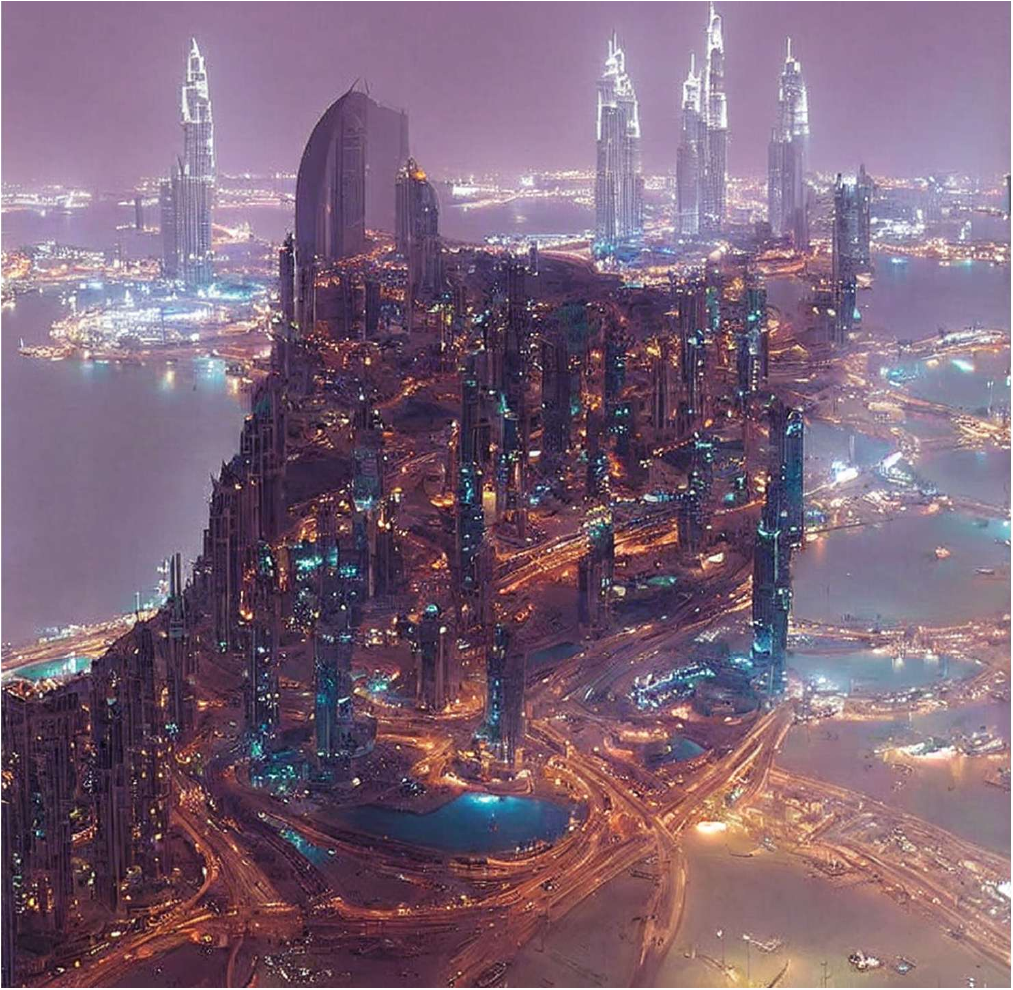}
\end{minipage}
\hspace{1em}
\begin{minipage}{.7\linewidth}
What is within this image? Is there any other content hidden within this image?

\textcolor{blue}{[The hidden object: a cat]}
\end{minipage}
\end{tcolorbox}

\paragraph{Follow-up hints.}
We also provide follow-up hints for the VLMs if the direct questions cannot get the answer we want.
For hidden text cases, we hint the model with:

\begin{tcolorbox}[fonttitle = \small\bfseries, title=
Follow-up Hint for Hidden Text,
colframe=gray!2!black,colback=gray!2!white,boxrule=1pt,boxsep=0pt,left=5pt,right=5pt,fontupper=\footnotesize, halign title = flush center]
Whether there is \texttt{[hidden text]} within this image?
\end{tcolorbox}

For hidden object cases, we hint:

\begin{tcolorbox}[fonttitle = \small\bfseries, title=
Follow-up Hint for Hidden Object,
colframe=gray!2!black,colback=gray!2!white,boxrule=1pt,boxsep=0pt,left=5pt,right=5pt,fontupper=\footnotesize, halign title = flush center]
Whether there is \texttt{[hidden figure or silhouette]} within this image?
\end{tcolorbox}

The \texttt{[hidden text]} is the correct answer text (e.g., ``POLO'') and \texttt{[hidden figure or silhouette]} is the brief description of the hidden object (e.g., ``a cat silhouette'').

\paragraph{Prompt engineering attempts.}
We should try explicit instructions for perceptual adjustments. For example, accompanied with the direct questions, we prompt the VLMs with this:

\begin{tcolorbox}[fonttitle = \small\bfseries, title=
Prompt Engineering Template,
colframe=gray!2!black,colback=gray!2!white,boxrule=1pt,boxsep=0pt,left=5pt,right=5pt,fontupper=\footnotesize, halign title = flush center]
Adjust contrast or brightness to examine the image macroscopically.
Zoom in or out to identify layered details.
\end{tcolorbox}

We should try to help the VLMs finish the work only by prompting.

\paragraph{Few-shot learning.}
Paired examples of original images, preprocessed versions (e.g., scaled or downsampled), and ground-truth answers should be input to the model to help it learn to understand and reproduce this process.


\subsection{SemVink Solutions}
\label{method:preprocess}


Like the cases shown in Figure~\ref{fig:problems}, the zero-shot prompting with both direct questions and follow-up hints fails to recognize hidden content. Therefore, we try preprocessing the image by scaling it like zooming out or adjusting their brightness or contrast like squint. As shown in Figure~\ref{fig:methods}, the two methods can help humans find the hidden content.

\paragraph{Squint.}
The squint method is also tested.
We keep the original image size and try different brightness and contrast adjustments. 
Moreover, we also try this enhancement on the image:
\textbf{Step 1.} Grayscale + Canny edge detection to highlight structural lines. \textbf{Step 2.} HSV color segmentation to isolate specific color regions. \textbf{Step 3.} Histogram equalization to improve contrast. The imaging result is provided for the model to help realize squinting automatically.




\paragraph{Zoom out.}
We implemented a preprocessing pipeline to simulate human-like perceptual adjustments.
For zoom-out operation, the input image are automatically resized to a lower resolution pixels (preserving the aspect ratio). The resized image is sent to the model with the original prompt to help the VLM have a zoomed-out view.

Our target is that the scaled images should be fed directly into VLMs without additional prompts. Therefore, we integrate the zoom-out and squint methods to the inference-time for each tested VLM.

\subsection{Embedding Redundancy}
\label{subsec:embedding}
We analyze vision encoder outputs for most models to understand failure modes.
To quantify feature redundancy in high-resolution embeddings, we extract vision encoder outputs (e.g., ViT-L/14) for both original and scaled-down images. Redundancy is measured through token repetition rate analysis which calculates the proportion of embedding tokens with cosine similarity >0.95 across spatial positions, indicating near-identical feature patterns. The attention map analysis on query tokens (e.g., ``\texttt{[HIDDEN\_TEXT]}'') using cross-attention layers shows that the attention across redundant regions (e.g., textures) in high-resolution images masks the activation on hidden content.

This methodology rigorously isolates VLMs’ limitations in low-level visual processing and demonstrates how simple preprocessing can bridge the gap between computational vision and human-like perception.

\section{Experiments}
\label{sec:experiments}

In this section, we present the performance of the SemVink zoom-out method integrated in VLMs. We conduct experiments by providing the VLM with each image in HC-Bench and direct questions. Follow-up hints will given if the direct questions cannot pass the test as described in Section~\ref{method:zeroshot}.
The comparison results validate the significant enhancement of SemVink and demonstrate that we find the way to let the models zoom.


\subsection{Experimental Setup}
\label{exp:setup}
The experiments are conducted on the constructed dataset HC-Bench as described in Section~\ref{subsec:data}.
We evaluate 12 state-of-the-art VLMs:
o3,\footnote{The model is available at \url{https://openai.com/index/introducing-o3-and-o4-mini/}}
o4-mini,\footnote{The model is available at \url{https://openai.com/index/introducing-o3-and-o4-mini/}}
Gemini 2.5 Pro,\footnote{The model is available at \url{https://deepmind.google/technologies/gemini/pro/}}
Grok 3,\footnote{The model is available at \url{https://grok3ai.net/}}
Mistral,\footnote{The model is available at \url{https://chat.mistral.ai/chat}}
Claude 3.7 Sonnet,\footnote{The model is available at \url{https://www.anthropic.com/claude/sonnet}}
LLaVA-v1.5-7B,\footnote{The model is available at \url{https://huggingface.co/liuhaotian/llava-v1.5-7b}}
Doubao-1.5-thinking-vision-pro,\footnote{The model is available at \url{https://www.volcengine.com/}}
Kimi-VL-A3B-Thinking,\footnote{The model is available at \url{https://huggingface.co/moonshotai/Kimi-VL-A3B-Thinking}}
Qwen2-VL-7B-Instruct,\footnote{The model is available at \url{https://huggingface.co/Qwen/Qwen2-VL-7B-Instruct}}
Qwen2-VL-72B-Instruct,\footnote{The model is available at \url{https://huggingface.co/Qwen/Qwen2-VL-72B-Instruct}}
and DeepSeek-VL2.\footnote{The model is available at \url{https://huggingface.co/deepseek-ai/deepseek-vl2}}

Accuracy (\%) for recognizing hidden text (exact match) and objects (category-level correctness) is calculated. Human evaluators manually verify responses to avoid ambiguities (e.g., partial matches or synonyms).
We define the correct answer of the text cases should exactly match the hidden word(s), but the object cases are deemed to take the recognition of the general category (e.g., ``face'' sufficed, no need for specific identity), considering that the knowledge across different models varies and our expectation is to check if the model can see any hidden content.



All the experiments are run on one NVIDIA A6000 GPU (48GB VRAM).




\begin{table}[t]
\centering
\resizebox{\linewidth}{!}{
\begin{tabular}{l | cccc}
Model
& 8–32
& 32–128
& 128–512
& 512+
\\
\hline
\claude
& \cmark & \cmark & \xmark & \xmark
\\
\gemini
& \xmark & \cmark & \xmark & \xmark
\\
\kimi
& \xmark & \cmark & \xmark & \xmark
\\
\qwx
& \cmark & \cmark & \xmark & \xmark
\\
\qwy
& \cmark & \cmark & \xmark & \xmark
\\
\end{tabular}
}
\caption{For one typical image containing text \textit{New York} as shown in Figure~\ref{fig:problems}, we test some models ability to recognize the hidden text by zooming out to different scales. We can find the range of the resolution from 32$\times$32 to 128$\times$128 (keep the aspect ratio) is the best zooming scale range.}
\label{tab:sense}
\end{table}
\begin{table}[t]
\centering
\resizebox{\linewidth}{!}{
\begin{tabular}{l | cccc}
Model
& B-32; C+32
& B-64; C+64
& B-128; C+64
& Enhance
\\
\hline
\claude
& \xmark & \xmark & \xmark & \xmark
\\
\gemini
& \xmark & \xmark & \xmark & \xmark
\\
\kimi
& \xmark & \xmark & \xmark & \xmark
\\
\qwx
& \xmark & \xmark & \xmark & \cmark
\\
\qwy
& \xmark & \xmark & \xmark & \xmark
\\
\end{tabular}
}
\caption{For one typical image containing text \textit{New York} as shown in Figure~\ref{fig:problems}, we test some models ability to recognize the hidden text by squinting. B-32; C+32 stands for brightness lowered by 32 and contrast enhanced by 32. No specific brightness, contrast or enhancement configuration can help the models.}
\label{tab:squint}
\end{table}

\begin{table*}[t]
\centering
\resizebox{\linewidth}{!}{
\begin{tabular}{l  ll ll ll ll ll}
\toprule
\multirow{2}{*}{Model}
& \multicolumn{2}{c}{Zero-Shot Direct}
& \multicolumn{2}{c}{Zero-Shot Hinted}
& \multicolumn{2}{c}{Zero-Shot Prompt}
& \multicolumn{2}{c}{Few-Shot}
& \multicolumn{2}{c}{w$\slash$ zoom-out}
\\
\cmidrule(lr){2-3}
\cmidrule(lr){4-5}
\cmidrule(lr){6-7}
\cmidrule(lr){8-9}
\cmidrule(lr){10-11}
& Text (\%)
& Object (\%)
& Text (\%)
& Object (\%)
& Text (\%)
& Object (\%)
& Text (\%)
& Object (\%)
& Text (\%)
& Object (\%)
\\
\midrule
\midrule
\othree
& 0 & 0 & 0 & 0
& 0 & 0 & 0 & 0
& \textbf{100.0}\textsubscript{\footnotesize \textcolor{green}{+100.0}}
& \textbf{100.0}\textsubscript{\footnotesize \textcolor{green}{+100.0}}
\\
\gpt
& 0 & 0 & 0 & 0
& 0 & 0 & 0 & 0
& \textbf{100.0}\textsubscript{\footnotesize \textcolor{green}{+100.0}}
& \textbf{100.0}\textsubscript{\footnotesize \textcolor{green}{+100.0}}
\\
\gemini
& 0 & 0 & 0 & 0
& 0 & 0 & 0 & 0
& \textbf{100.0}\textsubscript{\footnotesize \textcolor{green}{+100.0}}
& \textbf{100.0}\textsubscript{\footnotesize \textcolor{green}{+100.0}}
\\
\grok
& 0 & 5.36 & 0 & 8.93
& 0 & 5.36 & 0 & 5.36
& \textbf{98.21}\textsubscript{\footnotesize \textcolor{green}{+98.21}}
& \textbf{100.0}\textsubscript{\footnotesize \textcolor{green}{+91.07}}
\\
\mist
& 0 & 0 & 0 & 10.71
& 0 & 0 & 0 & 5.36
& \textbf{96.43}\textsubscript{\footnotesize \textcolor{green}{+96.43}}
& \textbf{100.0}\textsubscript{\footnotesize \textcolor{green}{+89.29}}
\\
\claude
& 0 & 0 & 1.78 & 3.57
& 0 & 0 & 0 & 0
& \textbf{98.21}\textsubscript{\footnotesize \textcolor{green}{+96.43}}
& \textbf{100.0}\textsubscript{\footnotesize \textcolor{green}{+96.43}}
\\
\llava
& 0 & 0 & 0 & 0
& 0 & 0 & 0 & 0
& \textbf{91.07}\textsubscript{\footnotesize \textcolor{green}{+91.07}}
& \textbf{98.21}\textsubscript{\footnotesize \textcolor{green}{+98.21}}
\\
\doubao
& 0 & 0 & 0 & 0
& 0 & 0 & 0 & 0
& \textbf{96.43}\textsubscript{\footnotesize \textcolor{green}{+96.43}}
& \textbf{98.21}\textsubscript{\footnotesize \textcolor{green}{+98.21}}
\\
\kimi
& 0 & 0 & 0 & 0
& 0 & 0 & 0 & 0
& \textbf{94.64}\textsubscript{\footnotesize \textcolor{green}{+94.64}}
& \textbf{91.07}\textsubscript{\footnotesize \textcolor{green}{+91.07}}
\\
\qwx
& 1.78 & 3.57 & 3.57 & 3.57
& 1.78 & 3.57 & 1.78 & 3.57
& \textbf{100.0}\textsubscript{\footnotesize \textcolor{green}{+96.43}}
& \textbf{96.43}\textsubscript{\footnotesize \textcolor{green}{+92.86}}
\\
\qwy
& 1.78 & 1.78 & 5.36 & 3.57
& 1.78 & 3.57 & 1.78 & 3.57
& \textbf{100.0}\textsubscript{\footnotesize \textcolor{green}{+94.64}}
& \textbf{100.0}\textsubscript{\footnotesize \textcolor{green}{+96.43}}
\\
\deepseek
& 0 & 0 & 0 & 0
& 0 & 0 & 0 & 0
& \textbf{92.86}\textsubscript{\footnotesize \textcolor{green}{+92.86}}
& \textbf{94.64}\textsubscript{\footnotesize \textcolor{green}{+94.64}}
\\
\bottomrule
\end{tabular}
}
\caption{The recognition accuracy across different VLMs with four methods mentioned in Section~\ref{method:zeroshot} and SemVink zoom-out method mentioned in Section~\ref{method:preprocess}. All tested VLMs are incapable of recognizing the hidden content in the images. With the help of SemVink zoom-out, each tested VLM obtains a nearly 100\% success rate.}
\label{tab:main}
\end{table*}

\begin{table*}[t]
\centering
\resizebox{\linewidth}{!}{
\begin{tabular}{l l l l l l}
\toprule
Model
& Zero Shot Direct (\%)
& Zero Shot Hinted (\%)
& Zero Shot Prompt (\%)
& Few Shot (\%)
& w$\slash$ zoom out (\%)
\\
\midrule
\midrule
\othree
& 0 & 0 & 0 & 0
& \textbf{98.11}\textsubscript{\footnotesize \textcolor{green}{+98.11}}
\\
\gpt
& 0 & 0 & 0 & 0
& \textbf{94.34}\textsubscript{\footnotesize \textcolor{green}{+94.34}}
\\
\gemini
& 0 & 0 & 0 & 0
& \textbf{90.57}\textsubscript{\footnotesize \textcolor{green}{+90.57}}
\\
\grok
& 0 & 1.89 & 0 & 0
& \textbf{92.45}\textsubscript{\footnotesize \textcolor{green}{+90.56}}
\\
\mist
& 0 & 3.77 & 1.89 & 0
& \textbf{94.34}\textsubscript{\footnotesize \textcolor{green}{+90.57}}
\\
\claude
& 0 & 1.89 & 0 & 0
& \textbf{98.11}\textsubscript{\footnotesize \textcolor{green}{+96.22}}
\\
\llava
& 0 & 0 & 0 & 0
& \textbf{96.23}\textsubscript{\footnotesize \textcolor{green}{+96.23}}
\\
\doubao
& 0 & 0 & 0 & 0
& \textbf{88.68}\textsubscript{\footnotesize \textcolor{green}{+88.68}}
\\
\kimi
& 0 & 0 & 0 & 0
& \textbf{86.79}\textsubscript{\footnotesize \textcolor{green}{+86.79}}
\\
\qwx
& 0 & 0 & 0 & 0
& \textbf{94.34}\textsubscript{\footnotesize \textcolor{green}{+94.34}}
\\
\qwy
& 0 & 0 & 0 & 0
& \textbf{96.23}\textsubscript{\footnotesize \textcolor{green}{+96.23}}
\\
\deepseek
& 0 & 0 & 0 & 0
& \textbf{84.90}\textsubscript{\footnotesize \textcolor{green}{+84.90}}
\\
\bottomrule
\end{tabular}
}
\caption{Validation of task difficulty on 53 internet-sourced hidden-content images, collected independently to reduce dataset-specific noise and biases.}
\label{tab:noise}
\end{table*}



\subsection{SemVink Evaluation}
\label{exp:preprocess}
We find zoom-out method is effective to help recognize the hidden content while squint fails.
We test some VLMs with different zoom-out scales and find the obvious sensitive range for VLMs to recognize the hidden information.
As shown in Table~\ref{tab:sense}, we find the best resolution is always in 32–128 pixels (keep the aspect ratio). Possible reason could be that higher resolutions reintroduce redundancy and lower resolutions degraded visibility.

Unlike the zoom-out method, we fail to obtain a good result with many squint configuration attempts. According to the method described in Section~\ref{method:preprocess}, we check some models with different brightness/contrast settings and the enhancement. As shown in Table~\ref{tab:squint}, squint operations are clearly insufficient to resolve the hidden-content recognition task.



According to the evaluation method in Section~\ref{method:zeroshot} and zoom-out in Section~\ref{method:preprocess}, we test all the twelve models with direct questions, hints after failing the direct question, prompt engineering, few-shot learning and zoom-out method. The experimental results are in Table~\ref{tab:main}.
Like the cases shown in Figure~\ref{fig:problems}, the results validate that all these methods but zoom-out lead to \textbf{catastrophic failures}.
Moreover, the prompt engineering for macroscopic view and few-shot learning method both hardly help VLMs. They even present worse performance than the hint method in zero-shot prompting.

Based on the experimental results and analysis above, we choose to conduct experiments to rigorously test the \textbf{zoom-out} method in the optimal range 32–128 pixels, so we employ the integration of our zoom-out method on all the tested VLMs and compare the results with the best results we obtained among methods of direct questions, hinted, prompt engineering and few-shot prompting, on all the 112 cases in HC-Bench.

According to the results in Table~\ref{tab:main},
we can find some remarkable patterns.

\paragraph{Universal failure of baseline methods.}
All VLMs achieve near-zero accuracy (0–5.36\%) on hidden text/object recognition under zero-shot, hinted, or few-shot settings.
Explicit instructions (e.g., ``zoom in/out to examine layered details'') yield no improvement, highlighting VLMs’ inability to simulate perceptual adjustments.

\paragraph{Dramatic improvement with zoom-out.}
Scaling images to low resolutions (32–128 pixels) achieves 91.07–100\% accuracy across models.
Larger models (e.g., \gpt, \gemini and \qwy) reach perfect scores of 100\% on both text and object cases, while smaller models (e.g., \kimi and \llava) also exceed 90\% accuracy overall.
Non-Latin text recognition (e.g., Chinese) improves proportionally, suggesting scaling generalizes across scripts.

\paragraph{Text vs. Object recognition.}
Hidden text cases have explicit character patterns amplified by scaling, while hidden object cases have category-level ambiguity (e.g., distinguishing Tyrannosaurus or dinosaurs resembled to other animals). Some models have a better performance in text cases while the others are better at object cases. A possible reason could be that different models have different preference in training data. As an overall pattern, the models cannot recognize either type of the hidden content without zoom-out.

\paragraph{Failure case analysis.}
Rare errors (1.79–8.93\%) occur due to two restricts. Severe artifacts: over-scaling merges critical details (e.g., thin strokes in Chinese characters).
Ambiguous object silhouettes: rare categories (e.g., Cologne Cathedral) lack distinct low-resolution patterns.
Also, encoder limitations matter. Smaller VLMs (e.g., LLaVA-7B) struggle with extreme downsampling due to limited receptive fields.

\subsection{Dataset-Specific Noise}
\label{exp:noise}
As the results shown in Table~\ref{tab:noise}, on the hidden-content images from wild on the internet, all tested VLMs still fail under standard settings, while the SemVink zoom-out method consistently yields large performance gains. This confirms that the limitation is inherent to current VLMs rather than an artifact of dataset-specific bias.



\begin{figure*}[t]
    \centering
    \includegraphics[width=\linewidth]{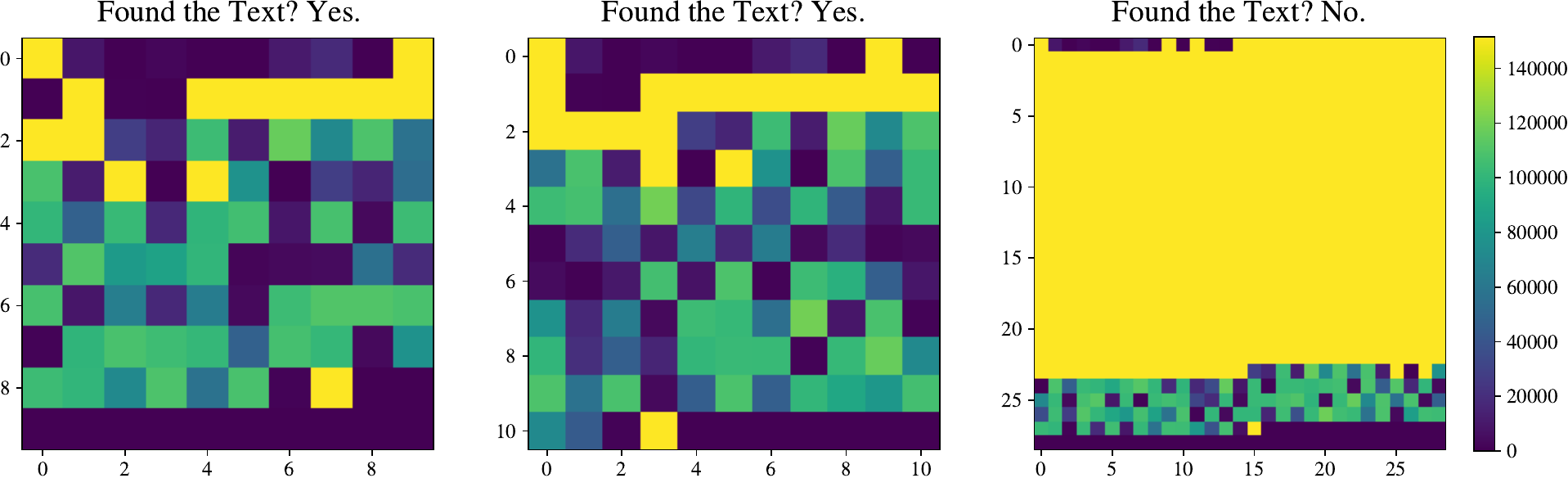}
    \caption{The visualization of the embeddings of the input prompts with the image. In the conditions of the left one (6 consecutive image tokens as in the consecutive \textcolor[rgb]{1,0.84,0.14}{yellow region} in the heatmap) and center one (10 consecutive image tokens), VLMs can recognize the hidden content. In the condition of the right one (666 consecutive image tokens), VLMs cannot find the hidden content. This demonstrates the redundant repeated information of the image is the key to obstruct finding the hidden content.}
    \label{fig:visualization}
    \vspace{-0.6em}
\end{figure*}

\subsection{Embedding Redundancy Analysis}
\label{exp:embed}

High-resolution images (512–1440 pixels) are with embedding tensors contained about 1000 repeated tokens which indicates redundant spatial patterns.
Scaled low-resolution images (32–128 pixels) are with a redundancy reduced to about 10 repeated tokens, aligning with successful detection.

In Figure~\ref{fig:visualization}, we visualize the 32-pixel scaled image, 128-pixel scaled image and 1024-pixel original image. We can find the clear patterns. The redundant features within the original image keep the VLMs from recognizing the hidden content.
Attention maps reveals that high-resolution embeddings focused excessively on background detailed information, masking hidden content.
Downsampled images shift attention to global structures, exposing hidden elements within the image.

Therefore, if we do not resize the image from a direct imaging degree but find and trim the relevant redundant part in embeddings, it is possible to integrate a general vision operation to VLMs.

\subsection{Discussion}
\label{exp:dis}


The failure to recognize hidden content in images exemplifies the lack of basic visual operations in current VLMs. As shown in Table~\ref{tab:squint}, only \qwx once successfully passed the squint test under the enhancement, while it fails in other cases and all other models consistently fail across all cases. VLMs struggle with tasks such as adjusting contrast and brightness and seeing from blurred visions, which are essential for recognizing patterns in blurry images, interpreting medical scans, and addressing security concerns. For instance, an adversarial trigger can be embedded in a seemingly normal image. While a general vision tool may not fully detect deeply hidden triggers, having such a tool as a foundational component is critical. Further technological advances can then build on this foundation to address security threats more effectively.

Such for medical imaging, VLMs currently reason based only on image and text tokens. Complex operations such as cropping or adjusting contrast in specific regions still depend on external software tools, akin to manual image editing before re-inputting into the VLM. The absence of integrated visual tools across existing VLMs leaves this area largely unexplored. However, our findings on the HC-Benchmark dataset suggest that such integration has the potential to significantly enhance overall performance.


\section{Conclusion}
\label{sec:conclusion}


This work reveals a critical limitation in vision-language models (VLMs). Current VLMs struggle to detect hidden content requiring human-like perceptual adjustments, as shown by their near-zero performance on our HC-Bench benchmark. This failure stems from prioritizing high-level semantics over low-level visual processing. Simple image scaling (32–128 pixels) resolves this limitation, achieving over 99\% accuracy by reducing redundant features in high-resolution embeddings. Our work exposes a critical flaw in VLM design and urges integration of multi-scale processing to bridge computational vision with human perceptual adaptability, advancing robustness in real-world vision-language applications.

\section*{Limitations}
\label{limit}


While our method demonstrates significant improvements, key limitations still remain: HC-Bench’s synthetic images may not fully capture real-world hidden content complexity, such as natural lighting or occlusion. The efficacy of programmatic scaling is resolution-dependent, potentially failing for ultra-fine patterns or requiring dynamic multi-scale sampling. Static downsampling neglects human-like dynamic adjustments (e.g., iterative zoom-contrast combinations), and rare scripts or categories may require specialized scaling thresholds. Computational costs for high-resolution preprocessing and energy trade-offs in scaling also warrant optimization. Finally, manual evaluation introduces subjectivity in object categorization, highlighting the need for automated metrics and adaptive multi-scale methods.

\section*{Acknowledgements}
The work is partially supported by the NSF of the United States Grant CRII 2451683, an NVIDIA Academic Grants Program, University of California at Merced, and a UC Merced Faculty Research Award.
The views and conclusions are those of the authors and should not reflect the official policy or position of the U.S. Government.

\bibliography{anthology,custom}
\bibliographystyle{acl_natbib}

\end{document}